\documentclass[11pt]{article}

\usepackage[final]{acl}

\usepackage{times}
\usepackage{latexsym}

\usepackage[T1]{fontenc}
\usepackage[utf8]{inputenc}
\usepackage{pifont}

\newcommand{\heart}{\ding{170}} % ♥
\newcommand{\spade}{\ding{171}} % ♠
\newcommand{\diamondshape}{\ding{169}} % ♦
\newcommand{\club}{\ding{168}} % ♣
\usepackage{microtype}

\usepackage{inconsolata}

\usepackage{graphicx}

\usepackage{lipsum}
\usepackage{booktabs}
\usepackage{multirow}
\usepackage{graphicx}
\usepackage{tabularx}
\usepackage{todonotes}
\usepackage{wrapfig}
\usepackage{amsmath}

\title{One-Topic-Doesn't-Fit-All: Transcreating Reading Comprehension Test \\for Personalized Learning}

\author{
  \textbf{Jieun Han\textsuperscript{\heart}},
  \textbf{Daniel Lee\textsuperscript{\spade}\thanks{Equal contribution.}},
  \textbf{Haneul Yoo\textsuperscript{\heart}\footnotemark[1]},
  \textbf{Jinsung Yoon\textsuperscript{\diamondshape}},
\\
  \textbf{Junyeong Park\textsuperscript{\heart}},
  \textbf{Suin Kim\textsuperscript{\club}},
  \textbf{So-Yeon Ahn\textsuperscript{\heart}},
  \textbf{Alice Oh\textsuperscript{\heart}}
\\[1em]
  \textsuperscript{\heart}KAIST \quad
  \textsuperscript{\spade}Salesforce AI Research \quad
  \textsuperscript{\diamondshape}Google Cloud \quad
  \textsuperscript{\club}Elice 
\\
\\
  \small{
    \textbf{Correspondence:} 
    \href{mailto:jieun_han@kaist.ac.kr}{jieun\_han@kaist.ac.kr}, 
    \href{mailto:ahnsoyeon@kaist.ac.kr}{ahnsoyeon@kaist.ac.kr}, 
    \href{mailto:alice.oh@kaist.edu}{alice.oh@kaist.edu}
  }
}

\begin{document}
\maketitle
\begin{abstract}
% The abstract should briefly summarize the contents of the paper in 150--250 words.
Personalized learning has gained attention in English as a Foreign Language (EFL) education, where engagement and motivation play crucial roles in reading comprehension. We propose a novel approach to generating personalized English reading comprehension tests tailored to students' interests. 
We develop a structured content \textit{transcreation} pipeline using OpenAI's gpt-4o, where we start with the RACE-C dataset, and generate new passages and multiple-choice reading comprehension questions that are linguistically similar to the original passages but semantically aligned with individual learners' interests.
Our methodology integrates topic extraction, question classification based on Bloom's taxonomy, linguistic feature analysis, and content transcreation to enhance student engagement. We conduct a controlled experiment with EFL learners in South Korea to examine the impact of interest-aligned reading materials on comprehension and motivation. Our results show students learning with personalized reading passages demonstrate improved comprehension and motivation retention compared to those learning with non-personalized materials.

\end{abstract}

\section{Introduction}
\begin{figure*}[tb!]
    \centering
    \includegraphics[width=\textwidth]{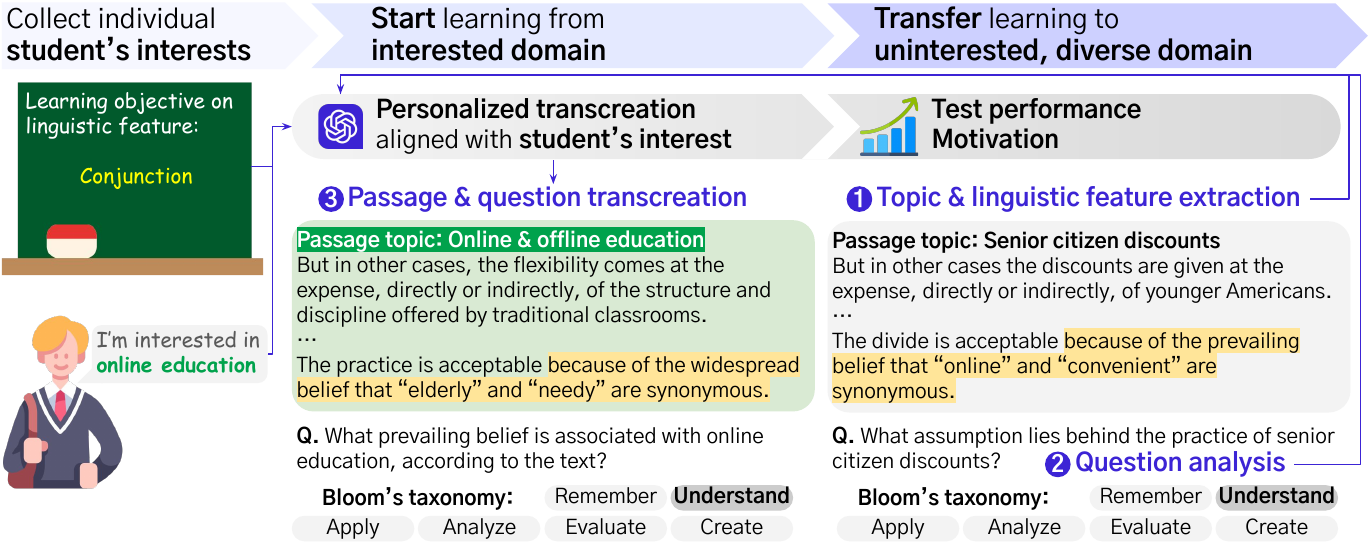}
    \caption{Transcreated reading passage and question. A source data with the uninterested topic (right) is transcreated into the student's interested topic (left).}
    \label{fig:main_figure}
    \vspace{-5mm}
\end{figure*}

%importance and benefits of personalized learning
Personalized learning has the potential to improve educational outcomes by offering learning content that meets the unique needs of individual learners \cite{bloom-1984-the,zhang-2020-understanding}. 
% However, despite its effectiveness, scaling personalized learning in real-world education remains impractical due to cost and logistical constraints.
% With advancements in large language models (LLMs), personalized learning has become increasingly feasible. 
% The advancements in large language models (LLMs) present a promising solution by enabling the automated generation of personalized learning materials, making adaptive education more accessible. 
% LLMs offer the potential to generate personalized learning materials that cater to individual student needs. 
% To effectively implement LLM-driven personalized learning, it is essential first to have a clear understanding of what personalized learning entails.
% Personalized learning is defined by four key factors: (1) cognitive---skills, knowledge, self-regulation; (2) affective---emotions, attitudes, regulation; (3) socio-cultural---context, identity, social agency; and (4) motivational---interest, goal orientation, and autonomy \cite{bernacki-2021-systematic,plass-2020-toward}.
Personalized learning is defined by four key aspects: (1) cognitive---encompassing skills, knowledge, and self-regulation; (2) affective---relating to emotions, attitudes, and regulation; (3) socio-cultural---shaped by context, identity, and social agency; and (4) motivational---driven by interest, goal orientation, and autonomy.
% Among these variables, our work focuses on motivational elements; in particular, topic interest has shown to have a positive impact on students' engagement, motivation, and knowledge acquirement across various educational contexts \cite{kahu2017student,alzubi2024students,lee2017impact}.
Our study focuses on motivational factors, particularly topic interest, which has been shown to positively influence students' engagement, motivation, and knowledge acquisition across various educational contexts \cite{kahu2017student,alzubi2024students,lee2017impact}.
% Among these, motivation plays a crucial role in sustaining engagement and improving knowledge retention \cite{yadav-2023-contextualizing}. 
% Interest-based personalization, in particular, has been found to enhance learning by increasing student motivation and making content more relevant. 
Personalized learning based on interest has been applied in structured subjects such as mathematics \cite{yadav-2023-contextualizing,hogheim-2015-supproting}, where concepts are largely universal and require minimal localization.  
Instead, English reading education presents greater adaptability through diverse reading materials that can be tailored to students' needs.
% On the other hand, English as a Foreign Language (EFL) learning heavily relies on semantic understanding and cultural context, making personalization even more crucial.
% In English as a Foreign Language (EFL) education, learners often struggle with motivation and comprehension due to limited exposure to authentic and personally relevant language input. 
% Unlike native speakers, who acquire language naturally through everyday interactions, EFL learners require engaging, context-aware materials to enhance retention and meaningful learning.
% Reading passages form the foundation of English reading education, but existing reading comprehension datasets do not align well with students' interests.
% current contents and tests are misaligned with students' interest
However, in EFL education, textbooks often present imbalanced content with Western-centric perspectives \cite{kang2009treating,lee-2009-situated,alawlaqi-2023-representation}.
This imbalance can lead to a decline in reading comprehension \cite{toti2022exploratory} or result in materials that are misaligned with the interests of non-Western EFL students.
% As an example, our analysis of the RACE-C \cite{liang-2019-a}, a multi-choice reading comprehension dataset collected from college English examinations, reveals that frequently appearing topics, such as Democratic Citizenship (14.4\%), are often perceived as uninteresting by students, with 22.5\% of students ranking it among their least engaging topics. 
% In contrast, highly engaging topics, such as Personal Life, are significantly underrepresented in the dataset (8.2\%). 
% This mismatch suggests that topics of current EFL reading materials are misaligned with EFL students interests, emphasizing the need for personalized content adaptation.
Despite the importance of personalized reading materials, most AI-driven educational content research has focused on question generation and assessment \cite{lu-2023-readingquizmaker,lamsiyah-2024-fine-tuning}, leaving a critical gap in the development of reading materials tailored to individual learners' interests.

% interest-based personalization content creation
Advancements in large language models (LLMs) offer a promising solution for creating personalized learning materials that were once deemed impractical due to cost and logistical constraints.
Specifically, a growing concept in NLP-based content adaptation is transcreation, which involves adapting content across cultural and contextual settings while preserving its intended meaning. 
Prior studies have primarily focused on cultural transcreation \cite{conia-etal-2024-towards,khanuja-etal-2024-image}, with limited attention to its broader educational applications. 
While Khanuja et al.~\cite{khanuja-etal-2024-image} introduced transcreation in math word problems, their modifications were surface-level, such as unit conversions, rather than substantive content adaptation.
% This research explores the impact of LLM-driven transcreation in enhancing engagement and learning outcomes in English reading education. 
By expanding transcreation to personalized learning, our study adapts reading materials to align with individual interests. 
% Specifically, we integrate reading material generation and question generation to deliver a more engaging and personalized learning experience for EFL students. 
We suggest that LLM-driven transcreation enhances comprehension and motivation retention in English reading education. 

\section{Method}

We suggest a personalized transcreation pipeline (\S\ref{sec:content_generation}) and examine the impact of tailored content on EFL learners (\S\ref{sec:student_experiment}).

\subsection{Personalized Transcreation}
\label{sec:content_generation}
We propose a systematic pipeline for generating personalized learning materials through LLM-driven transcreation. 
Using the RACE-C dataset as a source, our approach adapts reading passages and multiple-choice questions to align with individual student interests. 
The process is structured into five steps, each leveraging \texttt{gpt-4o} from OpenAI\thinspace\footnote{All following experiments using \texttt{gpt-4o} in this paper were conducted from Feb 1, 2025 to Feb 18, 2025 under OpenAI API services.} with an English teacher system prompt to ensure pedagogical quality.

\emph{\textbf{Step 1. Topic Extraction}}
We first analyze the topic of the reading passage in RACE-C. 
We incorporate nine categories and 33 subcategories from the educational curriculum by the Ministry of Education in South Korea \cite{moe-2022-ministry}. 
% These nine categories span personal, family, school, social life, culture, democratic citizenship, ecological transition, digital \& AI technologies, and general knowledge.

\emph{\textbf{Step 2. Question Analysis}}
Building on the prior research \cite{hwang-2024-towards,scaria-2024-automated}, we classify the question types from the original reading passage using six categories of Bloom's taxonomy \cite{bloom-1956-taxonomy}: \emph{Remember}, \emph{Understand}, \emph{Apply}, \emph{Analyze}, \emph{Evaluate}, and \emph{Create}. 
% Building on the research from \cite{hwang-2024-towards,scaria-2024-automated}, we generate relevant multiple-choice questions aligned with Bloom’s taxonomy levels. 

\emph{\textbf{Step 3. Linguistic Feature Extraction}}
In this step, we aim to identify and extract key linguistic features from the original reading passage.
We hypothesize that the sentences supporting the questions are essential for comprehending the text. 
Therefore, we employ 41 special tags to highlight these key linguistic features, as defined by the Ministry of Education in South Korea \cite{moe-2022-ministry}. 
We incorporate special tokens to be placed after sentences that support answering the questions for LLM to recognize important linguistic elements in Step 4.

\emph{\textbf{Step 4. Reading Passage Transcreation}}
We transcreate the original reading passage into a new version, adapting the topic to better align with student interests. 
When transcreating the passage in the extracted topic (Step 1) into the student's interested topic, we ensure consistency in the key linguistic features of the original passage (Step 3).
% We provide the original passage and annotate it with special tokens on linguistics features following each sentence that supports a comprehension question.
% This method allows the language model to identify and integrate key linguistic features from the original passage into the new version.

\emph{\textbf{Step 5. Question Transcreation}}
We transcreate questions, options, and answers using the question type information of the original question set from Step 2. 
Since the main content of the reading passage changed, the corresponding questions and answer choices must also be adapted accordingly. 
By maintaining alignment with the original question types, we ensure that the revised questions reflect the same cognitive demands and assessment items.

\emph{\textbf{Validation}}
After transcreation, we have an English expert validate the reading passage and question sets. 
This expert holds Secondary School Teacher's Certificates (Grade II) for English, licensed by the Ministry of Education, Republic of Korea. 
The expert ensures that each question is answerable based on the given passage. 
Among the transcreated questions, 2.8\% were deemed unanswerable, all stemming from a single question that required understanding the temporal relationship between two events.
This issue arises because current LLMs struggle with temporal reasoning, which relies on multiple cognitive skills \cite{xiong-etal-2024-large}. 
To address this, the expert modified the reading passages when necessary to make the questions answerable. 
On average, 1.7 words were added per modified passage, all of which were time-related, including years and temporal conjunctions.

\begin{table}[htb!]
\centering
\vspace{1mm}
\begin{tabular}{@{}l|c|c|c@{}}
\toprule
                                            & \multicolumn{2}{c|}{\textbf{Test 1}} & \textbf{Test 2} \\ \cmidrule(lr){2-3}
                                            & \textbf{Group A} & \textbf{Group B} &                  \\ \midrule
\# words                                    & $\text{387}_{\pm\text{94.68}}$ & $\text{372.45}_{\pm\text{66.55}}$ & $\text{394}_{\pm\text{78.35}}$ \\
TTR                                         & $\text{0.59}_{\pm\text{0.05}}$    & $\text{0.59}_{\pm\text{0.04}}$    & $\text{0.52}_{\pm\text{0.45}}$   \\
FRES               & $\text{31.24}_{\pm\text{16.86}}$  & $\text{34.02}_{\pm\text{11.39}}$  & $\text{52.20}_{\pm\text{11.36}}$ \\ \bottomrule
\end{tabular}
\caption{Analysis on reading passages of each test}
\label{tab:article_analysis}
\end{table}

We employ an automated approach to further evaluate the quality and difficulty of transcreated reading passages and questions. 
Specifically, we use the LLM-as-a-Judge method \cite{zheng-2024-judging} to assess whether the transcreated questions align with the same cognitive level in Bloom’s taxonomy. 
The evaluation results indicate an average accuracy of 0.72 and a Cohen’s Kappa score of 0.29, suggesting a fair level of agreement.
Table~\ref{tab:article_analysis} shows linguistic features and difficulty level of the reading passages, measured by the word counts, type-token ratio (TTR), and Flesch Reading Ease Score (FRES)~\cite{flesch1979how}.
% We ensure consistent difficulty and linguistic features of the reading passages across tests
% We ensure consistent difficulty and linguistic features of the reading passages across tests.
% Specifically, the average word count for the main test in Group A, Group B, and Test 2 was $\text{387.00}_{\pm\text{94.68}}$, $\text{372.45}_{\pm\text{66.55}}$, and $\text{394.0}_{\pm\text{78.35}}$ words on average, respectively.
% Lexical variety, measured by type-token ratio (TTR), is $\text{0.59}_{\pm\text{0.05}}$ (Group A), $\text{0.59}_{\pm\text{0.04}}$ (Group B), and $\text{0.52}_{\pm\text{0.45}}$ (Test 2).
% We use the Flesch Reading Ease Score (FRES)~\cite{flesch1979how} to assess readability.
% FRES for the main test are $\text{31.24}_{\pm\text{16.86}}$ (Group A) and $\text{34.02}_{\pm\text{11.39}}$ (Group B), indicating a college-level difficulty (U.S. grade level $\text{14.25}_{\pm\text{1.78}}$ and $\text{13.69}_{\pm\text{2.18}}$, respectively~\cite{kincaid1975derivation}).
% Test 2 is slightly more readable, with a FRES of $\text{52.20}_{\pm\text{11.36}}$, corresponding to grade $\text{11.26}_{\pm\text{2.75}}$.

\subsection{Student Experiment}
\label{sec:student_experiment} 

We recruit EFL learners from a college in South Korea, selecting participants with scores from internationally standardized English proficiency tests such as TOEFL.
The student cohort comprises 20 Korean students with a gender distribution of 10 females and 10 males. 
To ensure a balanced level of English proficiency across groups, we divide the students into two groups of 10  while maintaining a similar distribution of TOEFL scores. The average TOEFL score for Group A is $\text{92.15}_{\pm\text{14.95}}$, and for Group B is $\text{93.13}_{\pm\text{12.09}}$.

% Our experiment consists of two reading comprehension tests: the transcreated test (Test 1) and the original test sourced from RACE-C (Test 2). 
Students first take the transcreated test (Test 1) and then take the original test sourced from RACE-C (Test 2). 
We assess their comprehension and motivation retention with uninteresting topics in Test 2 compared to Test 1.
This design is based on the idea that exposure to diverse topics is essential for EFL students \cite{little-2022-broadening}, as focusing solely on personally interesting topics may hinder overall knowledge expansion.
Each test comprises four passages with five questions each, totaling 100 points.
Before the test, we collect students' individual interests in 33 topics using a 7-point Likert scale. 
Additionally, students provide open-ended responses identifying their top four most interesting topics, while their least interesting topics are selected through a multiple-choice format from the 33 given topics.
% Based on these responses, Group A is assigned reading materials that are transcreated with randomly selected topics, whereas Group B receives transcreated reading materials aligned with their interests.
Group A receives transcreated reading materials with randomly selected topics, while Group B is given transcreated reading materials aligned with their stated interests. 
Specifically, a single original text from the RACE-C dataset is transcreated into randomly selected topics for Group A and into each student’s preferred topic for Group B.
For the source test, we select four less preferred passages from the RACE-C, guided by a preliminary survey: 2.b (family events, holidays, household chores, and daily family life), 5.a (cultural differences among generations and genders), 4.c (social events---meetings, community events, graduations, weddings, and funerals), and 6.b (democratic and global citizenship---human rights, gender equality, global etiquette, and peace)
% The topic ranked as the most uninteresting---9.c (patriotism, peace, security, education on Dokdo, and unification)---is excluded due to the absence of relevant passages in RACE-C. 
% We select the following four topics: 2.b (family events, holidays, household chores, and daily family life), 5.a (cultural differences among generations and genders), 4.c (social events---meetings, community events, graduations, weddings, and funerals), and 6.b (democratic and global citizenship---human rights, gender equality, global etiquette, and peace).

After completing each test, students are given time to review and reflect on the given task. 
This review session allows them to analyze their mistakes and reinforce their understanding of the content. 
Additionally, the student submits open-ended answers on their learning experience. 
Regarding the written reflection, the study employed a thematic approach to examine patterns of engagement, comprehension, and motivation across two test phases. 
This allowed for a comparative assessment of how personalized versus non-personalized reading materials influenced learners' experiences.
% Participant responses were analyzed to identify emerging themes and changes over time, allowing for a comparative interpretation of how personalized and non-personalized reading materials influenced learners' experiences.
% This review session allows them to analyze their mistakes and reinforce their understanding of the content. 
% Students first take the Test 1 (transcreated) and then take the Test 2 (original source), which assesses their ability to apply the knowledge and skills gained from the initial test. 
Students complete the Instructional Materials Motivation Survey (IMMS) \cite{keller-2010-motivational} on a 7-point Likert scale after Test 1 and Test 2.

\section{Experimental Result}
We present quantitative results on comprehension and motivation and qualitatively analyze students' open-ended responses to their test impressions.

\begin{figure}[tb!]
    \centering
    \includegraphics[width=0.6\columnwidth]{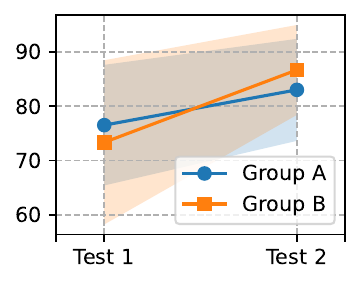}
    \vspace{-5mm}
    \caption{Test score differences within Group A/B}
    \label{fig:total_score}
    \vspace{-5mm}
\end{figure}

\emph{\textbf{Quantitative Result}}
All the following significance tests are conducted with a $p$ value of 0.01.
% Figure~\ref{fig:total_score} presents the test scores for each group, comprising 20 questions, totaling 100 points.
Figure~\ref{fig:total_score} shows group test scores for 20 questions, totaling 100 points.
The Wilcoxon Signed-Rank test demonstrates that Group B, which received personalized reading materials, achieves a significant improvement in performance ($\text{73.33}_{\pm\text{15.12}} \rightarrow \text{86.67}_{\pm\text{8.28}}$).
In contrast, Group A, given randomly assigned reading materials, does not exhibit a significant difference ($\text{76.50}_{\pm\text{11.11}} \rightarrow \text{83.00}_{\pm\text{9.39}}$).
% In particular, Group B reports significant improvements in the \emph{Analyze} questions  compared to Group A ($+\text{1.33}_{\pm\text{0.71}} \rightarrow \text{0.00}_{\pm\text{0.45}}$), out of five points on average. 
In particular, Group B reports significant improvements in the \emph{Analyze} questions ($+\text{1.33}_{\pm\text{0.71}}$) compared to Group A ($+\text{0.00}_{\pm\text{0.45}}$), out of five points on average.
This suggests that Group B excels in high-level analytical questions, as such questions above the \emph{Analyze} level are rare in the source dataset (5\%).
Furthermore, Group B significantly reduces their test turnaround time after reading personalized materials ($\text{30.78}_{\pm\text{7.60}} \rightarrow \text{27.78}_{\pm\text{5.30}}$ minutes), compared to Group A ($\text{24.50}_{\pm\text{11.04}} \rightarrow \text{32.00}_{\pm\text{17.77}}$ minutes).
% In particular, Group B reports significant improvements in the \emph{Analyze} questions  compared to Group A ($+\text{1.33}_{\pm\text{0.71}} \rightarrow \text{0.00}_{\pm\text{0.45}}$), out of five points on average.
% These results suggest that providing personalized reading materials aligned with students' interests enhances reading comprehension in second language education.

It also correlates with IMMS, which captures students' motivational perceptions.
Across all groups, Test 2, containing less engaging topics, is perceived as less motivating.
However, the Mann-Whitney U test confirms that Group B shows better motivation retention ($\text{4.68}_{\pm\text{0.66}} \rightarrow \text{4.61}_{\pm\text{1.04}}$) than Group A ($\text{4.82}_{\pm\text{0.38}} \rightarrow \text{4.44}_{\pm\text{0.85}}$) on a 7 Likert scale.
Specifically, Group B reports increased confidence as they encounter similarly structured articles ($\text{4.30}_{\pm\text{0.66}} \rightarrow \text{4.57}_{\pm\text{1.19}}$), while Group A becomes less confident as topics change ($\text{5.01}_{\pm\text{0.82}} \rightarrow \text{4.30}_{\pm\text{0.66}}$).
These results further reinforce the efficacy of personalized reading materials in enhancing engagement in EFL education.

\emph{\textbf{Qualitative Result}}
In the written responses, Group B participants demonstrate sustained engagement and consistent performance, even as external motivational factors diminish. 
In contrast, Group A exhibits a deepening disengagement over time, highlighting the cumulative impact of non-personalized content on learner motivation. 
Participants in Group A, who receive non-personalized materials, consistently report low engagement. 
In the initial phase, A1 and A2 mention that they often solve problems without thoroughly reading the passages. 
This lack of deep engagement may contribute to reduced concentration, as indicated by A3, who experiences a significant decline in focus as the test progresses. 
Over time, this disengagement becomes more pronounced. After Test 2, A2 and A4 explicitly describe the reading tasks as unengaging, attributing their disinterest to the repetitive structure of the passages. 
This suggests that a lack of personalized content not only results in immediate disengagement but may also lead to a cumulative decline in motivation across tasks.

In contrast, participants in Group B, who initially receive interest-aligned reading materials, exhibit higher engagement. 
Early responses indicate that four participants (B1, B2, B3, B4) found the readings well-matched to their interests, leading to greater satisfaction (B2) and enhanced comprehension (B4).
This alignment appears to foster an initial sense of motivation and involvement. 
However, engagement in Group B was not entirely stable. 
While interest-aligned content generally enhances motivation, some participants (B4, B5) experience a mismatch between their expectations and the actual content. 
For example, B5’s initial interest in tennis leads to a passage about younger generations' enthusiasm for the sport, but they later realize that their true interest lies in improving tennis skills rather than broader discussions about the topic. 
This suggests that interest-based content is most effective when it precisely reflects the learner’s expectations, rather than broadly aligning with a general topic of interest. 
Despite the removal of interest-aligned content in the later phase, participants in Group B (B3, B6, B7) remain engaged. 
% B3, B6, and B7 continue to find the reading tasks interesting, even without direct alignment with their personal preferences. 
% Additionally, B8, B9, and B2 find the second test easier due to their familiarity with the problem structure, suggesting that prior experience with similar formats contributes to a sense of ease in task completion.

\section{Conclusion}
Our study demonstrates that LLM-driven personalized transcreation significantly enhances comprehension in English reading education by aligning reading materials with individual student interests. 
% The results indicate that students who received interest-aligned reading passages perform better and report higher motivation compared to those given randomly assigned materials. 
The initial motivation from personalized content can carry over, helping learners sustain engagement even when motivation-related factors (interest-aligned content) are removed.
% Future research should explore more fine-grained personalization, focusing on deeper interest alignment to mitigate misalignment issues.
Future research should explore a fine-grained interest alignment to mitigate misalignment and evaluate longitudinal outcomes for the sustained impact.
We also suggest expanding LLM-driven personalized learning to incorporate cultural and cognitive factors.
% culture-algined transcreation
% Beyond interest-aligned personalization, our findings highlight the importance of cultural transcreation in ensuring content relevance. 
% U.S.-centric references, such as Obama in our experiment, may hinder engagement for EFL learners in South Korea. 
% Replacing such elements with culturally familiar figures maintains contextual appropriateness and fosters a deeper connection with the material. 
% Additionally, cultural transcreation extends beyond simple entity replacement, enabling the adaptation of historical narratives and societal contexts to align with learners' cultural frameworks.
% Future research should refine contextual adaptation techniques and explore the long-term impact of transcreated materials on learning retention and critical thinking. Expanding cultural transcreation for multilingual and cross-cultural education could further enhance the adaptability of AI-driven learning materials, making them more inclusive and effective across diverse educational settings.

\section*{Acknowledgment}
This work was supported by Institute for Information \& communications Technology Promotion(IITP) grant funded by the Korea government (MSIP) (No. RS-2024-00443251, Accurate and Safe Multimodal, Multilingual Personalized AI Tutors). This research was supported by the MSIT(Ministry of Science, ICT), Korea, under the Global Research Support Program in the Digital Field program(RS-2024-00436680) supervised by the IITP(Institute for Information \& Communications Technology Planning \& Evaluation). This project is supported by Microsoft Research Asia.

\bibliography{custom}

\end{document}